\documentclass[runningheads,a4paper]{llncs}

\usepackage{amssymb}
\usepackage{graphicx}
\usepackage{multirow}  

\usepackage{url}
\urldef{\mailsa}\path|dfsousa@lasige.di.fc.ul.pt|
\newcommand{\keywords}[1]{\par\addvspace\baselineskip
\noindent\keywordname\enspace\ignorespaces#1}

\usepackage{breakcites}  
\usepackage{hyperref}

\begin{document}

\mainmatter  

\title{Using Neural Networks for Relation Extraction from Biomedical Literature}

\titlerunning{Neural Networks for Biomedical Relation Extraction}

%
%
\author{Diana Sousa\thanks{dfsousa@lasige.di.fc.ul.pt}%
\and Andre Lamurias\and Francisco M. Couto}
\authorrunning{Neural Networks for Biomedical Relation Extraction}

\institute{LASIGE, Faculdade de Ci\^{e}ncias, Universidade de Lisboa, Portugal\\}

%
%

\maketitle

\begin{abstract}
Using different sources of information to support automated extracting of relations between biomedical concepts contributes to the development of our understanding of biological systems. The primary comprehensive source of these relations is biomedical literature. Several relation extraction approaches have been proposed to identify relations between concepts in biomedical literature, namely, using neural networks algorithms. The use of multichannel architectures composed of multiple data representations, as in deep neural networks, is leading to state-of-the-art results. The right combination of data representations can eventually lead us to even higher evaluation scores in relation extraction tasks. Thus, biomedical ontologies play a fundamental role by providing semantic and ancestry information about an entity. The incorporation of biomedical ontologies has already been proved to enhance previous state-of-the-art results.

\keywords{Relation Extraction, Biomedical Literature, Neural Networks, Deep Learning, Ontologies, External Sources of Knowledge}
\end{abstract}

\section{Introduction}

Biomedical literature is the main medium that researchers use to share their findings, mainly in the form of articles, patents, and other types of written reports \cite{1}. A researcher working on a specific topic needs to be up-to-date with all developments regarding the work done on the same topic. However, the volume of textual information available widely surpasses the ability of analysis by a researcher, even if restricting it to a domain-specific topic. Not only that, but the textual information available is usually in an unstructured or highly heterogeneous format. Thus, retrieving relevant information requires not only a considerable amount of manual effort but is also a time-consuming task.

Scientific articles are the primary source of knowledge for biomedical entities and their relations. These entities include human phenotypes, genes, proteins, chemicals, diseases, and other biomedical entities inserted in specific domains. A comprehensive source for articles on this topic is the PubMed platform \cite{2}, combining over 29 million citations while providing access to their metadata. Processing this volume of information is only feasible by using text mining solutions.

Automatic methods for information extraction (IE) aim at obtaining useful information from large data-sets \cite{3}. Text mining uses IE methods to process text documents. Text mining systems usually include named-entity recognition (NER), named-entity linking (NEL), and relation extraction (RE) tasks. NER consists of recognizing entities mentioned in the text by identifying the offset of its first and last character. NEL consists of mapping the recognized entities to entries in a given knowledge base. RE consists of identifying relations between the entities mentioned in a given document. 

RE can be performed by different methods, namely, by order of complexity, co-occurrence, pattern-based (manually and automatically created), rule-based (manually and automatically created), and machine learning (feature-based, kernel-based, and recurrent neural networks (RNN)).

In recent years, deep learning techniques, such as RNN, have proved to achieve outstanding results at various natural language processing (NLP) tasks, among them RE. The success of deep learning for biomedical NLP is in part due to the development of word vector models like Word2Vec \cite{4}, and, more recently, ELMo \cite{5}, BERT \cite{6}, GPT \cite{7}, Transformer-XL \cite{8}, and GPT-2 \cite{9}.  These models learn word vector representations that capture the syntactic and semantic word relationships and are known as word embeddings. Long Short-Term Memory (LSTM) RNN constitute a variant of artificial neural networks presented as an alternative to regular RNN \cite{10}.  LSTM networks deal with more complex sentences, making them more fitting for biomedical literature. To improve their results in a given domain, it is possible to integrate external sources of knowledge such as domain-specific ontologies.

\begin{figure}[ht]
\centering
\includegraphics[width=9cm]{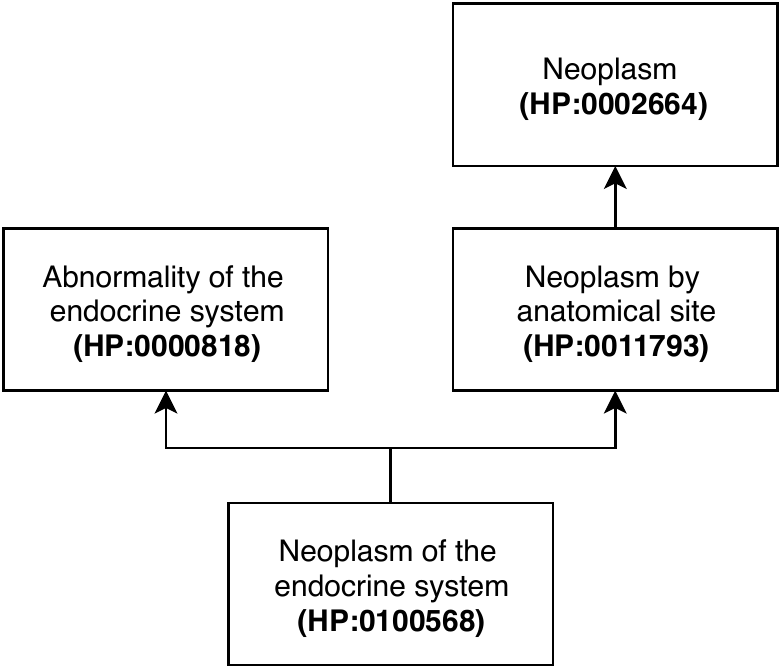}
\fontsize{9}{10.8}\caption[HPO Ontology Excerpt]{An excerpt of the HPO ontology showing the first ancestors of \textit{neoplasm of the endocrine system}, using \textbf{is-a} relationships}
\label{figure:1}
\end{figure}

The knowledge encoded in the various domain-specific ontologies, such as the Gene Ontology (GO) \cite{11}, the Chemical Entities of Biological Interest (ChEBI) ontology \cite{12}, and the Human Phenotype Ontology (HPO) \cite{13}  is deeply valuable for detection and classification of relations between different biomedical entities. Besides that, these ontologies make available important characteristics about each entity; they also provide us with the underlying semantics of the relations between those entities, such as is-a relations. For example, \textit{neoplasm of the endocrine system} (HP:0100568), a phenotypic abnormality that describes a tumor (abnormal growth of tissue) of the endocrine system \textbf{is-a} \textit{abnormality of the endocrine system} (HP:0000818), and \textbf{is-a} \textit{neoplasm by anatomical site} (HP:0011793), which in turn \textbf{is-a} \textit{neoplasm} (HP:0002664) (Fig. \ref{figure:1}). 

The information provided by the ancestors is not expressed directly in the text and can support or disprove an identified relation. Ontologies are formally organized in machinereadable formats, facilitating their integration in relation extraction models. 

Using different sources of information, as additional data, to support automating searching for relations between biomedical concepts contributes to the development of pharmacogenomics, clinical trial screening, and adverse drug reaction identification \cite{14}. Identifying new relations can help validate the results of recent research, and even propose new experimental hypotheses.

\section{Related Work}

\noindent{This section presents the basic concepts and resources that support Relation Extraction (RE) deep learning techniques, namely, Natural Language Processing (NLP), Text Mining Primary Tasks, and
initial approaches for Relation Extraction.}


\hypertarget{2.1}{\subsection{Natural Language Processing}}

Natural language processing (NLP) is an area in computer science that aims to derive meaning from unstructured or highly heterogeneous text written by humans. NLP covers several techniques that constitute preprocessing steps for the tasks described in Subheading \hyperlink{2.2}{2.2}. These NLP techniques have different goals and are often combined to obtain higher performance.

\begin{itemize}

\item{\textbf{Tokenization}: has the purpose of breaking the text into tokens to be processed individually or as a sequence. These tokens are usually words but can also be phrases, numbers and other types of elements. The most straightforward form of tokenization is breaking the input text by the use of whitespace or punctuation. However, with scientific biomedical literature, that is usually descriptive and formal, we have to account for complex entities like human phenotype terms (composed of multiple words), genes (represented by symbols), and other types of structured entities. These entities tend to be morphologically complex and need specialized tokenization pipelines. Some researchers use a compression algorithm \cite{15}, byte pair encoding (BPE), to account for biomedical vocabulary variability. BPE represents open vocabularies through a fixed-size vocabulary of variablelength character sequences, making it suitable for neural networks models.} 

\item{\textbf{Stemming and Lemmatization}: aims at reducing the variability of natural language by normalizing a token to its base form (stem) \cite{16}. It can also take into account the context of the
token, along with vocabulary and morphological analysis to determine the canonical form of the word (lemma). The stem can correspond only to a fragment of a word, but the lemma is always a real word. For instance, the stem of the word \textit{having} is \textit{hav} and the lemma is \textit{have}.}

\item{\textbf{Part-of-Speech Tagging}: consists of assigning each word of a sentence to the category where it belongs taking into account their context (e.g., verb or preposition). Each word can belong to more than one category. This feature is useful to gain information on the role of a word in a given sentence.}

\item{\textbf{Parse Tree}: represents the syntactic structure of a sentence. There are two different types of parse trees: constituency-based parse trees and dependency-based parse trees. The main difference between the two is that the first distinguishes between the terminal and non-terminal nodes and the second does not (all nodes are terminal). In constituency-based parse trees, each node of the tree is either a \textit{root} node, a \textit{branch} node, or a \textit{leaf} node. For each given sentence there is only one \textit{root} node. The \textit{branch} node connects to two or more \textit{child} nodes, and the \textit{leaf} node is terminal. These leaves correspond to the lexical tokens \cite{17}. Dependency-based parse trees are usually simpler because they only identify the primary syntactic structure, leading to fewer nodes. Parse trees generate structures that are used as inputs for other algorithms and can be constructed based on supervised learning techniques.}

\end{itemize}


\hypertarget{2.2}{\subsection{Text Mining Primary Tasks}}

Text mining has become a widespread approach to identify and extract information from unstructured or highly heterogeneous text \cite{18}. Text mining is used to extract facts and relationships in a structured form that can be used to annotate specialized databases and to transfer knowledge between domains \cite{19}. We may consider text mining as a sub-field of data mining. Thus, data mining algorithms can be applied if we transform text to a proper data representation, namely, numeric vectors. Even if in recent years text mining tools have evolved considerably in number and quality, there are still many challenges in applying text mining to scientific biomedical literature. The main challenges are the complexity and heterogeneity of the written resources, which make the retrieval of relevant information, that is, relations between entities, a nontrivial task.

Text Mining tools can target different tasks together or separately. Some of the primary tasks are Named Entity Recognition (NER), Named-Entity Linking (NEL) and Relation Extraction (RE).

\begin{itemize}

\item{\textbf{Named Entity Recognition (NER)}:  seeks to recognize and classify entities mentioned in the text by identifying the offset of its first and last character. The workflow of this task starts by splitting the text in tokens and then labeling them into categories (part-of-speech (POS) tagging).}

\item{\textbf{Named-Entity Linking (NEL)}: maps the recognized entities to entries in a given knowledge base. For instance, a gene can be written in multiple ways and mentioned by different names or acronyms in a text. NEL links all these different nomenclatures to one unique identifier. There are several organizations dedicated to providing identifiers, among them the National Center for Biotechnology Information (NCBI) for genes, and the Human Phenotype Ontology (HPO) for phenotypic abnormalities encountered in human diseases.}

\item{\textbf{Relation Extraction (RE)}: identifies relations between entities (recognized manually or by NER) in a text. Tools mainly consider relations by the co-occurrence of the entities in the same sentence, but some progress is being made to extend this task to the full document (taking into account a global context) \cite{20}.}

\end{itemize}

\begin{figure}[hbt!]
\centering
\includegraphics[width=9cm]{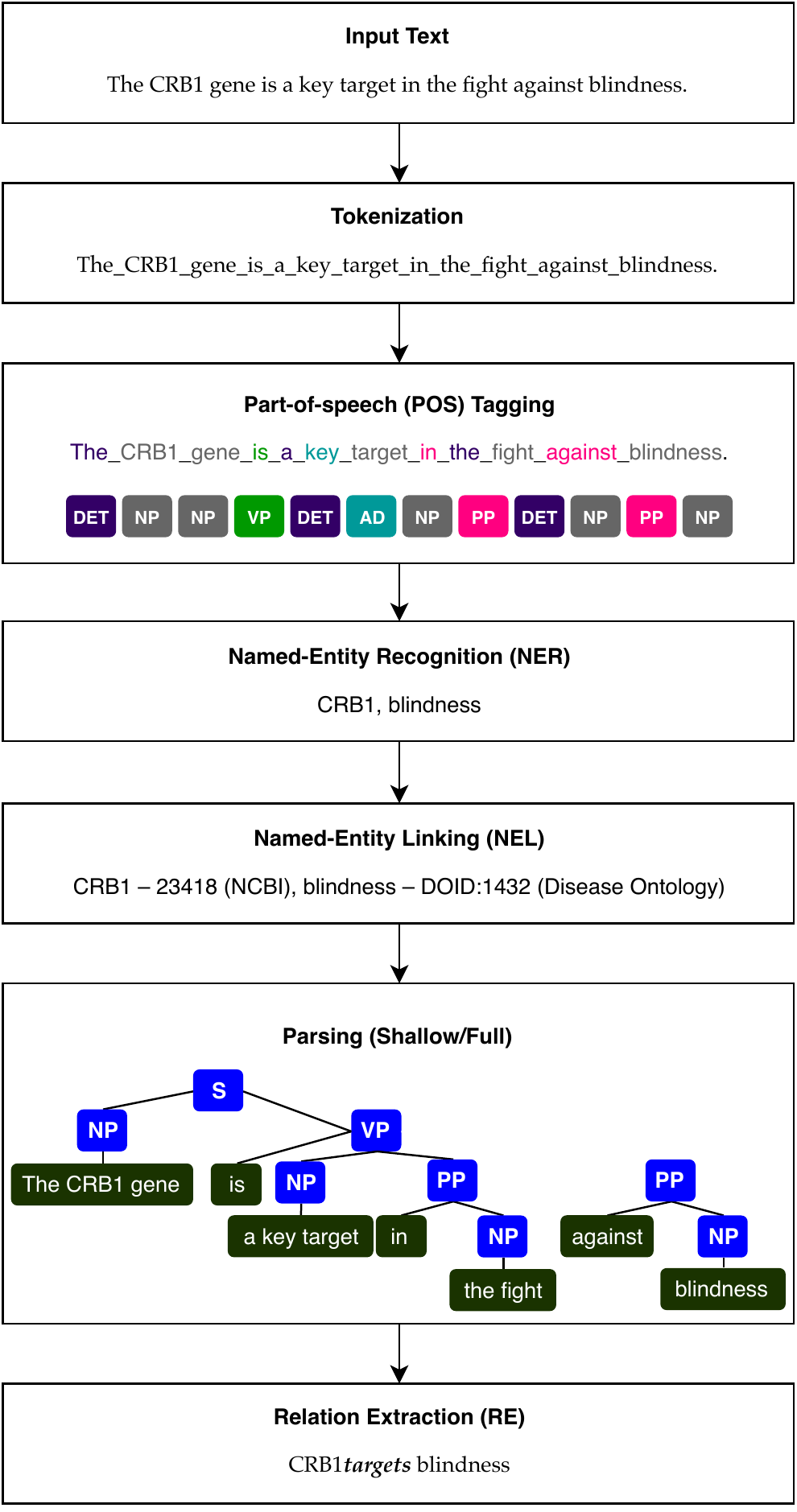}
\fontsize{9}{10.8}\caption[Relation Extraction Workflow]{Workflow of a simplified RE system. (Text obtained from \cite{21})}
\label{figure:2}
\end{figure}

The workflow of a typical RE system is presented in Fig. \ref{figure:2}.

\subsection{Initial Approaches for Relation Extraction}

Through the years, several approaches have been proposed to extract relations from biomedical literature \cite{22}. Most of these approaches work on a sentence level to perform RE, due to the inherent complexity of biomedical literature.

\begin{itemize}

\item{\textbf{Co-occurrence}: assumes that if two entities are mentioned in the same sentence (co-occur), it is likely that they are related. Usually, the application of this approach results in a higher recall (most of the entities co-occurring in a sentence participate in a relation), and lower precision. Some methods use frequency-based scoring schemes to eliminate relations identified by chance \cite{23}. Nowadays, most applications use co-occurrence as a baseline against more complex approaches \cite{24}.}

\item{\textbf{Pattern-based}: uses manually defined patterns and automatically generated patterns to extract relations. \textbf{Manually defined patterns} require domain expertise knowledge about the type of biomedical entities, their interactions, and the text subject at hand. Initial systems made use of regular expressions to match word patterns that reflected a relation between two entities \cite{25}, making use of a dictionary of words that express relation, such as \textit{trigger} and \textit{stimulate}.  Later systems introduce part-of-speech when applied to complex sentences, such as the ones that we typically find in biomedical literature \cite{26}. Opposite to the co-occurrence approaches, manually defined patterns frequently achieve high precision but tend to have poor recall. This approach does not generalize well, and therefore is difficult to apply to new unseen data. \textbf{Automatically generated patterns} encompass two main approaches, bootstrapping with \textit{seeds} \cite{27} and leveraging of the corpora \cite{28}. The bootstrapping method uses a small set of relations known as \textit{seeds} (e.g., gene-disease pairs). The first step is to identify the \textit{seeds} in the data-set and map the relation pattern they describe. The second step is to try to apply the mapped patterns to the data-set to identify new pairs of relations that follow the same construction. Finally, the original set of relations is expanded by adding these new pairs. When after repeating all previous steps no more pairs are found, the process ends. Some systems apply distant supervision techniques to keep track of the validity of the added patterns. Distant supervision uses existing knowledge base entries as gold standards to confirm or discard a relation. This method is susceptible to noisy patterns, as the original set of relations grows. On the other hand, the leveraging of the corpora method makes immediately use of the entire data-set to generate the patterns. This
method requires a higher number of annotated relations and produces highly specific patterns that are unable to match new unseen data. Automatically generated patterns can achieve a higher recall than manually defined patterns, but overall the noisy patterns continue to damage the precision. Nevertheless, there are a few efforts to reduce the number of noisy patterns \cite{29}.}

\item{\textbf{Rule-based}: also uses manually defined and automatically generated rules from the training data to extract relations. Depending on the systems, the differences between pattern-based and rule-based approaches can be minor. Ruled-based approaches not only use patterns but also additional restraints to cover issues that are difficult to express by patterns, such as checking for the negation of the relations \cite{30}. Some ruled-based systems distance themselves from pattern-based approaches by replacing regular expressions with heuristic algorithms and sets of procedures \cite{31}. Like pattern-based approaches, ruled-based approaches tend to have poor recall, even though rules tend to be more flexible. The trade-off recall/precision can be improved using automatic methods for rule creation \cite{32}.}

\item{\textbf{Machine Learning (ML)-based}: usually makes use of large annotated biomedical corpora (supervised learning) to perform RE. These corpora are preprocessed using NLP tools and then used to train classification models. Beyond Neural Networks, described in detail in Subheading \hyperlink{3}{3}, it is possible to categorize ML methods into two main approaches, Feature-based and Kernel-based. \textbf{Feature-based approaches} represent each instance (e.g., sentence) as a vector in an \textit{n}-dimensional space. Support Vector Machines (SVM) classifiers tend to be used to solve problems of binary classification, and are considered \textit{black-boxs} because there is no interference of the user in the classification process. These classifiers can use different features that are meant to represent the data characteristics (e.g., shortest path, bag-of-words (BOW), and POS tagging) \cite{33}. \textbf{Kernel-based approaches} main idea is to quantify the similarity between the different instances in a data-set by computing the similarities of their representations \cite{34}. Kernel-based approaches add the structural representation of instances (e.g., by using parse trees). These methods can use one kernel or a combination of kernels (e.g., graph, sub-tree (ST), and shallow linguistic (SL)).}

\end{itemize}


\hypertarget{3}{\section{Neural Networks for Relation Extraction}}

Artificial neural networks have multiple different architectures implementations and variants. Often data representations are used as added sources of information to perform text mining tasks, and ontologies may even be used as external sources of information to enrich the model.


\subsection{Architectures}

\begin{figure}[hbt!]
\centering
\includegraphics[width=10cm]{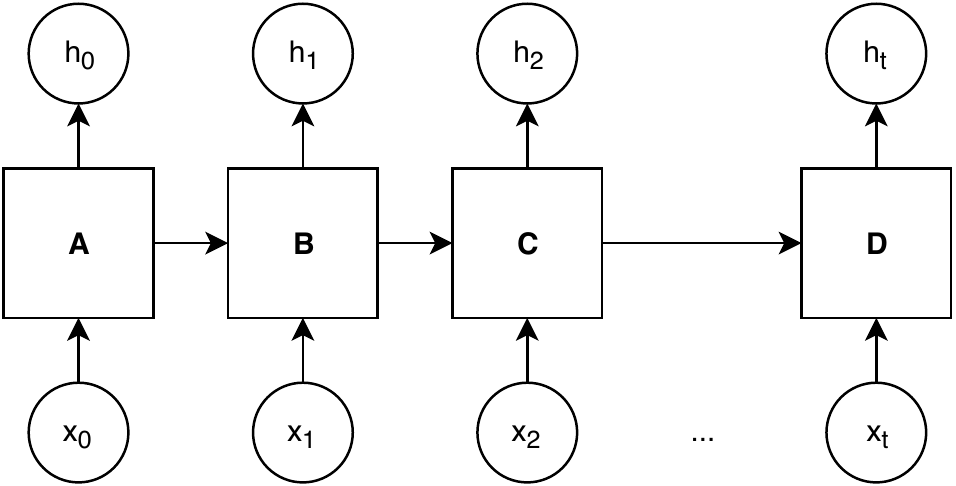}
\fontsize{9}{10.8}\caption[Artificial Neural Networks Architecture]{Architecture representation of an artificial neural networks model, where x\textsubscript{0-t} represents the inputs and h\textsubscript{0-t} the respective outputs, for each module from A to D}
\label{figure:3}
\end{figure}

\textbf{Artificial Neural Networks} are a parallel combination of small processing units (nodes) which can acquire knowledge from the environment through a learning process and store the knowledge in
the connections between the nodes \cite{35} (represented by direct graphs \cite{36}) (Fig. \ref{figure:3}). The process is inspired by the biological brain function, having each node correspond to a \textit{neuron} and the connections between the nodes representing the \textit{synapses}.

\begin{figure}[hbt!]
\centering
\includegraphics[width=10cm]{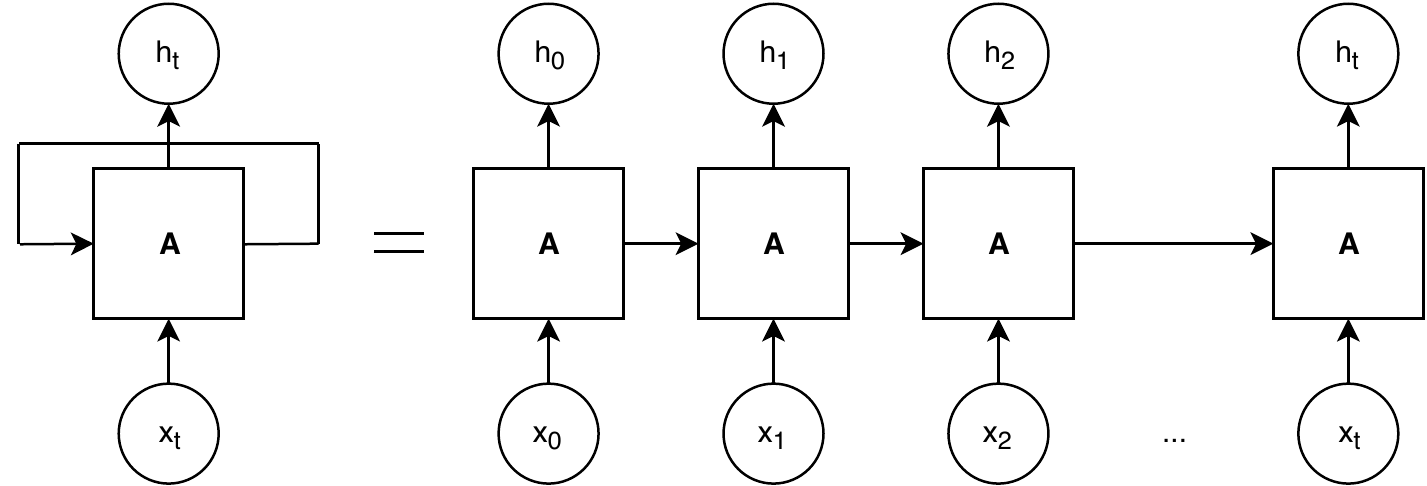}
\fontsize{9}{10.8}\caption[Recurrent Neural Networks Architecture]{Architecture representation of a recurrent neural networks model, where x\textsubscript{0-t} represents the inputs and h\textsubscript{0-t} the respective outputs, for the repeating module A}
\label{figure:4}
\end{figure}

\textbf{Recurrent Neural Networks} (RNN) is a type of artificial neural network where the connections between the nodes are able to follow a temporal sequence. This means that RNN can use their internal state, or \textit{memory}, to process each input sequence (Fig. \ref{figure:4}). Deep learning techniques, such as RNN, aim to train classification models based on word embeddings, part-of-speech (POS) tagging, and other features. RNN classifiers have multilayer architectures, where each layer learns a different representation of the input data. This characteristic makes RNN classifiers flexible for application to multiple text mining tasks, without requiring task-specific feature engineering.

\begin{figure}[hbt!]
\centering
\includegraphics[width=10cm]{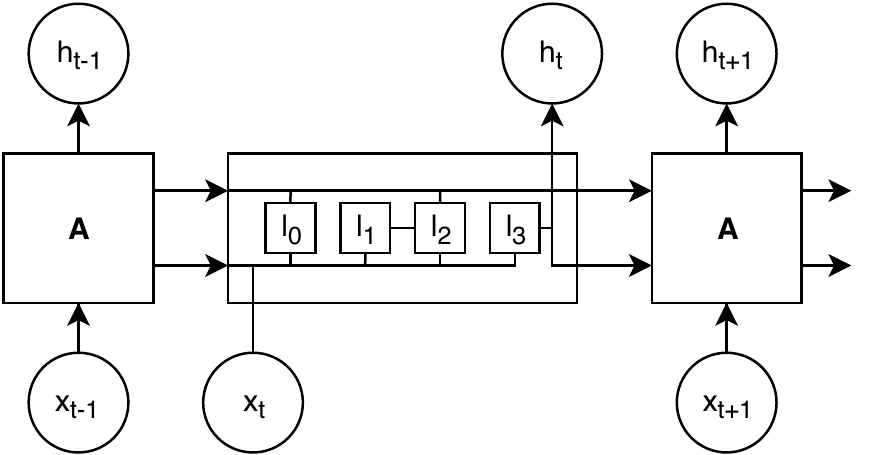}
\fontsize{9}{10.8}\caption[Long Short-Term Memory Networks Architecture]{Architecture representation of a long-short-term memory networks model, where x\textsubscript{0-t} represents the inputs and h\textsubscript{0-t} the respective outputs, for the repeating module A, where each repeating module has four interacting layers (l\textsubscript{0-3})}
\label{figure:5}
\end{figure}

\textbf{Long Short-Term Memory} (LSTM) networks are an alternative to regular RNN \cite{10}. LSTMs are a type of RNN that handles long dependencies (e.g., sentences), making this classifier more suitable for the biomedical domain, where sentences are usually long and descriptive (Fig. \ref{figure:5}). In recent years, the use of LSTMs to perform Relation Extraction (RE) tasks has become widespread in various domains, such as semantic relations between nominals \cite{37}. \textbf{Bidirectional LSTMs} use two LSTM layers, at each step, one that reads the sentence from right to left, and other that reads from left to right. The combined output of both layers produces a final score for each step. Bidirectional LSTMs has yield better results than traditional LSTMs when applied to the same data-sets \cite{38}.

\hypertarget{3.2}{\subsection{Data Representations}}

The combination of multiple and different language and entityrelated data representations is vital for the success of neural network models dedicated to RE tasks. Some of these features were already described in Subheading \hyperlink{2.1}{2.1}, such as POS tagging and parse trees.

\textbf{Shortest Dependency Path (SDP)} is a feature that identifies the words between two entities mentioned in the text, concentrating the most relevant information while decreasing noise \cite{39}. 

\textbf{Word Embeddings} are fixed-sized numerical vectors that aim to capture the syntactic and semantic word relationships. These word vectors models use multiple different pretraining sources, for instance, Word2Vec \cite{4} uses English Wikipedia, and BERT \cite{6} uses both English Wikipedia and BooksCorpus. Early models, such as Word2Vec, learned one representation per word, but this proved to be problematic due to polysemous and homonymous words. Recently, most systems started to apply one embedding per word sense. One of the reasons why BERT outperforms previous methods is because it uses contextual models, meaning that it generates a unique representation for each word in a sentence. For instance, in the sentences fragments, \textit{they got \textbf{engaged}}, and \textit{students were very \textbf{engaged} in}, the word \textit{engaged} for non-contextual models would have the same meaning. BERT also outperforms other word vector models that take into account the sentence context, such as ELMo \cite{5} and ULMFit \cite{40}, due to being an unsupervised and deeply bidirectional pre-trained language representation. 

\textbf{WordNet Hypernyms} are a feature that helps to hierarchize entities, structuring words similar to direct acyclic graphs \cite{41}. For example, \textit{vegetable} is a hypernym of \textit{tubers}, which in turn constitutes a hyponym of \textit{vegetable}. This feature is comparable to an ontology in the sense that a hierarchy relation is identified, but is missing the identification of the relations between the different terms. 

Using different features as information sources feeding individual channels leads to multichannel architecture models. Multichannel approaches were already proven to be effective in RE tasks \cite{39}. 

Regarding biomedical RE, LSTMs were successful in identifying drug-drug interactions \cite{42}, gene-mutation relations \cite{43}, drug-mutation relations \cite{44}, among others. Some methods use domain-specific biomedical resources to train features for biomedical tasks. BioBERT \cite{45} is a domain specific language representation model pre-trained on large-scale biomedical corpora, based on BERT \cite{6} architecture. BioBERT, using minimal task-specific architecture modifications, significantly outperforms previous biomedical state-of-the-art models in the text mining primary tasks of Named-Entity Recognition, Named-Entity Linking, and RE. The BR-LSTM \cite{46} model uses a multichannel approach with pre-trained medical concept embeddings. Using the Unified Medical Language System (UMLS) concepts, BR-LSTM applies a medical concept embedding method developed by Vine et al. \cite{47}. BO-LSTM \cite{48} uses the relations provided by domain-specific ontologies to aid the identification and classification of relations between biomedical entities in biomedical literature. 


\subsection{Ontologies}

An ontology is a structured way of providing a common vocabulary in which shared knowledge is represented \cite{49}. Word embeddings can learn how to detect relations between entities but manifest difficulties in grasping the semantics of each entity and their specific domain. Domain-specific ontologies provide and formalize this knowledge. Biomedical ontologies are usually structured as a directed acyclic graph, where each node corresponds to an entity and the edges correspond to known relations between those entities. Thus, a structured representation of the semantics between entities and their relations, an ontology, allows us to use it as an added feature to a machine learning classifier. Some of the biomedical entities structured in publicly available ontologies are genes properties/attributes (gene ontology (GO)), phenotypes (human phenotype ontology (HPO)), diseases (disease ontology (DO)), and chemicals (chemical entities of biological interest (ChEBI)). All of these entities participate in relations with different and same domain type entities. Hence, the information about each entity on a semantic level adds a new layer of knowledge to increase the performance of RE classifiers.

Non-biomedical models using ontologies as an added source of information to neural networks is becoming widespread for several tasks. Li et al. \cite{50} propose using word sense definitions, provided by the WordNet ontology, to learn one embedding per word sense for word sense disambiguation tasks. Ma et al. \cite{51} focus their work on semantic relations between ontologies and documents, using the DBpedia ontology. Some researchers explored graph embedding techniques \cite{52}  that convert relations to a low dimensional space which represents the structure and properties of the graph. Other researchers have combined different sources of information, including ontological information, to do multi-label classification \cite{53} and used ontology concepts to represent word tokens \cite{54}.

However, few authors have used biomedical ontologies to perform RE. Textpresso \cite{55}  is a text-mining system that works as a search engine of individual sentences, acquired from the full text of scientific articles, and articles themselves. It integrates biomedical ontological information (e.g., of genes, phenotypes, and proteins) allowing for article and sentence search query by term. The integration of the ontological information allows for semantic queries. This system helps database curation by automatically extracting biomedical relations. The IICE \cite{56}  system uses kernel-based support vector machines along with an ensemble classifier to identify and classify drug–drug interactions, linking each chemical compound to the ChEBI ontology. Tripodi et al. \cite{57}  system focus on drug–gene/protein interaction discovery to aid database curation, making use of ChEBI and GO ontologies. BO-LSTM \cite{48}  is the only model until now that incorporates ancestry information from biomedical ontologies with deep learning to extract relations from the text, specifically drug–drug interactions and gene–phenotype relations. 


\hypertarget{4}{\section{Evaluation Measures}}

The evaluation of machine learning systems is done by applying the trained models to a gold standard test-set, manually curated or annotated by domain experts and unseen by the system. For a Relation Extraction (RE) task, the gold standard test-set should correspond to the list of pairs of entities (e.g., phenotype–gene or gene–disease pairs) that co-occur in the same sentences and their relation (\textit{Known} or \textit{Unknown}).

To any given information extraction system it is necessary to define what constitutes a positive and negative result. In RE tasks the types of results possible are shown in Table \ref{table:evaluation}.

\begin{table}[!ht]
\renewcommand\arraystretch{1.2}
\small
\caption[Types of Results Obtained with an Information Extraction System for a RE Task]{Types of results obtained with an information extraction system for a RE task}
\centering
\begin{tabular}{ |c|c|c| }
\hline
\textbf{Annotator (Gold Standard)} & \textbf{System} & \textbf{Classification}\\
\hline\hline
\multirow{2}{*}{Relation} & Relation & True Positive (TP) \\
\cline{2-3}
 & No Relation & False Negative (FN) \\ 
\hline
\multirow{2}{*}{No Relation} & Relation & False Positive (FP) \\
\cline{2-3}
 & No Relation & True Negative (TN) \\
 \hline
\end{tabular}
\label{table:evaluation}
\end{table}

The primary goal of a given information retrieval system is to maximize the number of TP (True Positives) and TN (True Negatives). To compare results obtained with different data-sets or different tools we have three distinct evaluation metrics: recall, precision and F-measure. Precision represents how often the results are correct, recall is the number of correct results identified and F-measure is a combination of both metrics that expresses overall performance, being the harmonic mean of precision and recall:

\begin{equation}
\scriptsize
Recall = \frac{TP}{TP + FN}
\qquad
Precision = \frac{TP}{TP + FP}
\qquad
F-measure = \frac{2\times Precision\times Recall}{Precision + Recall}
\label{equation:evaluation}
\end{equation}

\vspace{0.3cm}

The performance of the most recent systems dedicated to biomedical RE, described in Subheading \hyperlink{3.2}{3.2}, is shown in Table \ref{table:evaluation_soa}. These systems are not comparable, since each system is focused on the relations between different biomedical entities, and even addresses more than binary relations, such as the Graph LSTM (GOLD) system.

\begin{table}[!ht]
\renewcommand\arraystretch{1.2}
\small
\caption[Biomedical RE Systems Performance]{Biomedical RE systems current performance}
\centering
\begin{tabular}{ |c|c|c|c| }
\hline
\textbf{System} & \textbf{Precision} & \textbf{Recall} & \textbf{F-Measure}\\
\hline\hline
DLSTM \cite{42} & 0.7253 & 0.7149 & 0.7200 \\
\hline
Graph LSTM (GOLD) \cite{43} & 0.4330 & 0.3050 & 0.3580 \\ 
\hline
BioBERT \cite{45} & 0.8582 & 0.8640 & 0.8604 \\
\hline
BR-LSTM \cite{46} & 0.7152 & 0.7079 & 0.7115 \\
\hline
BO-LSTM \cite{48} & 0.6572 & 0.8184 & 0.7290 \\
\hline
\end{tabular}
\label{table:evaluation_soa}
\end{table}

For RE tasks a human acceptable performance is usually around 85/90\% in F-measure  \cite{58}. To facilitate the creation of gold standards we should strive for semi-automation, that is, employ automatic methods for corpora annotation (creating silver standard
corpora), and then correct those annotations using domain-specific curators.


\hypertarget{5}{\section{Conclusions}}

The way we communicate scientific knowledge is through scientific literature. At the current rate of document growth, the only way to
process this amount of information is by using computational methods. The information obtained through these methods can lead to a better understanding of biological systems. When creating biomedical text mining systems, it is essential to take into account
the specific characteristics of biomedical literature. Biological information follows different nomenclatures and levels of sentence
complexity.

Automatic biomedical Relation Extraction (RE) still has a long way to go to achieve human-level performance scores. Over recent
years, some innovative systems have successively achieved better results by making use of multiple knowledge sources and data
representations. These systems not only rely on the training data but make use of different language and entity-related features, to
create models that identify relations in highly heterogeneous text. This chapter described the evolution of dedicated RE systems, up until neural networks. Neural networks, especially deep neural networks, which make use of multichannel architectures are composed of multiple features. Each system integrates different features distinctively. An optimal combination of features and the ideal
features to perform biomedical RE tasks are still far from human level performance.

The knowledge encoded in biomedical ontologies plays a vital part in the development of deep neural networks systems, providing semantic and ancestry information for entities, such as genes, proteins, phenotypes, and diseases. One could also add semantic similarity measures as an additional information layer. Also, to accompany the growth of systems targeting different biomedical relations, there is a growing need for more domain-specific corpora, that can only be accomplished by automated corpus creation (silver standard corpora) \cite{59}.

Integrating different knowledge sources instead of relying solely on the training data for creating classification models will
allow us not only to find relevant information for a particular problem quicker, but also to validate the results of recent research,
and propose new experimental hypotheses.

\end{document}